# STATISTICAL TUNING OF ARTIFICIAL NEURAL NETWORK

⚘ **Mohamad Yamen AL Mohamad** * [1], ⬤ **Hossein Bevrani**† [1]**, and** ⬤ **Ali Akbar Haydari**‡[1]

[1]Faculty of Mathematics, Statistics, and Computer Sciences, University of Tabriz, Tabriz, Iran

## ABSTRACT

Neural networks are often regarded as "black boxes" due to their complex functions and numerous parameters, which poses significant challenges for interpretability. This study addresses these challenges by introducing methods to enhance the understanding of neural networks, focusing specifically on models with a single hidden layer. We establish a theoretical framework by demonstrating that the neural network estimator can be interpreted as a nonparametric regression model. Building on this foundation, we propose statistical tests to assess the significance of input neurons and introduce algorithms for dimensionality reduction, including clustering and (PCA), to simplify the network and improve its interpretability and accuracy. The key contributions of this study include the development of a bootstrapping technique for evaluating artificial neural network (ANN) performance, applying statistical tests and logistic regression to analyze hidden neurons, and the assessing neuron efficiency. We also investigate the behavior of individual hidden neurons in relation to output neurons and apply these methodologies to the IDC and Iris datasets to validate their practical utility. This research advances the field of Explainable Artificial Intelligence by presenting robust statistical frameworks for interpreting neural networks, thereby facilitating a clearer understanding of the relationships between inputs, outputs, and individual network components.

## 1 Introduction

Deep learning has achieved remarkable success across a broad spectrum of application domains, ranging from economics [31] and medical science [30] to social network analysis [2], bioinformatics [14], and healthcare [25]. These successes are largely due to the ability of deep learning models to capture and model complex patterns and relationships that traditional methods struggle to handle. However, despite its impressive performance, deep learning is often criticized for its "black-box" nature, where the internal workings of the models remain opaque [21]. This lack of interpretability poses challenges, particularly in critical fields such as healthcare and autonomous systems, where understanding how a decision is made is essential for trust and accountability. In addition to interpretability issues, deep learning models demand extensive computational resources, large amounts of labeled data, and long training times, often consuming significant energy during the process [21]. A typical deep learning model, such as a convolutional neural network, involves optimizing millions of parameters, abstracting features from data that may not directly correspond to physical realities. While these hidden features are vital for tasks like classification and prediction, they further obscure the models decision-making process. The demand for explainable models has grown significantly, especially in high-stakes applications where transparency is as crucial as accuracy [7]. For example, in autonomous driving, understanding why a vehicle made a particular decision is essential, particularly in the event of an accident. DNNs are complex systems, often compared to Russian nesting dolls, with layers encapsulating numerous mathematical functions that makes it difficult to interpret the models and understand how they arrive at specific outcomes. To address these challenges, the field of Explainable Artificial Intelligence (XAI) has emerged, aiming to develop methods that make deep learning models more interpretable and transparent [3]. XAI focuses on explaining the results and the structure of models according to predefined standards, balancing the need for interpretability with model accuracy. As deep learning continues to influence critical areas of society, the need for interpretability has become not only a scientific but also a legal and social concern, prompting calls for more transparent AI systems [26].

---
*yamenmohamad1401@ms.tabrizu.ac.ir

†Corsponding auther, bevrani@tabrizu.ac.ir

‡heydari@tabrizu.ac.ir

This report explores the current limitations of deep learning models in terms of interpretability and computational burden and examines the emerging solutions provided by XAI to enhance understanding and trust in artificial intelligence systems. This research aims to provide a comprehensive statistical explanation of deep learning models by leveraging statistical methods. Specifically, it seeks to develop statistical models capable of interpreting the components of pre-trained deep networks, offering a novel approach that balances traditional AI methods with statistical explanations. The research addresses key questions related to the use of statistical models to simulate, interpret, and explain the behavior and relationships within deep networks. While traditional DNNs prioritize accuracy, their interpretability remains a significant challenge. Existing methods typically rely on post hoc techniques to make sense of network decisions, often resulting in approximations rather than full explanations of the decision-making process. Furthermore, these approaches can yield conflicting interpretations, highlighting the need for a more robust methodology. Though some statistical approaches, such as Bayesian Neural Networks and Probabilistic Neural Networks, have been explored, they do not fully address the interpretation of conventional deep learning models. This research aims to fill this gap by introducing statistical models that provide a clearer, more consistent explanation of DNNs internal mechanisms. DNNs have primarily focused on optimizing prediction accuracy, but the interpretability of their decisions remains a critical challenge. Typically, the interpretation of these models relies on post hoc techniques, which aim to explain the models behavior after it has been trained [21]. These methods are often used to understand high-level features and the overall decision-making process. However, since interpretability is considered only after the network's architecture is selected and trained, this process often provides approximate explanations that may lack clarity and coherence [26]. Moreover, different post hoc models can lead to varying, and sometimes conflicting, interpretations of the same net- works decisions, undermining the reliability of these methods. Despite the widespread use of DNNs, there has been limited work on interpreting traditional deep learning models using statistical approaches. However, some alternative methodologies grounded in statistical principles have been explored. For instance, Medeiros et al. [15] introduced an artificial neural network framework built on statistical methodologies, particularly targeting time series analysis. Their approach views network components as random variables, offering a statistical foundation for interpreting the models. Similarly, Bayesian ANNs [13] incorporate prior distributions on network weights and update them with posterior distributions after training, embedding probabilistic reasoning into the network. This statistical approach improves the interpretability of neural networks by introducing a probabilistic framework.

Another notable development is Specht's Neural Probabilistic Network [27], which uses probability density functions within feedforward neural networks. This method is widely applied in classification and prediction tasks, offering a more interpretable structure by using probabilistic principles. While these statistical approaches offer valuable insights, most are designed for specific applications and architectural designs, such as regression networks [28]. However, they do not seamlessly extend to interpreting conventional pre-trained DNNs. As such, while they offer potential pathways for enhancing interpretability, they do not yet provide comprehensive solutions for explaining deep learning models in general. Bootstrap methodologies have been widely explored for constructing (ANNs and DNNs). Notable works in this area include studies by Sharma and Tiwari [22], Reed et al. [18], Secchi et al. [24], Huang et al. [10], and Chillotti et al. [5]. Among these, the method proposed by Michelucci and Venturini [16] is particularly significant for its innovative approach to bootstrap ANN construction.

Despite these advancements, recent research has cast doubt on the practical utility of bootstrap methods in deep learning. In their study, Nixon et al. [17] from Google Research's Brain Team demonstrated that bootstrap methods may not be as beneficial as previously believed. Their findings suggest that ensemble methods where each model is independently trained on a bootstrapped version of the dataset have consistently achieved state-of-the-art results in predictive accuracy, uncertainty estimation, and robustness to out-of-distribution data. Although bootstrapping is well-established in decision tree literature and frequentist statistics, it is infrequently applied in practice for DNNs.

A central hypothesis for the limited success of bootstrap methods in deep learning involves the percentage of unique data points in the subsampled dataset. Even after adjustments for this factor, bootstrap ensembles of DNNs do not show significant advantages over simpler baseline models. This raises important questions about the effectiveness of data randomization techniques in enhancing deep learning models.

Several important contributions for evaluating neural network predictors through bootstrapping are noteworthy. Weigend and LeBaron [29] investigated the use of bootstrapping to assess neural network predictors, providing foundational insights into this approach. Carney et al. [4] explored confidence and prediction intervals for neural network ensembles, offering practical methods for incorporating bootstrapping into model evaluation. Didona and Romano [6] analyzed bootstrapping techniques for machine learning performance prediction via analytical models, contributing to a deeper understanding of this methodology. The statistical tuning plays a crucial role in optimizing and understanding model performance. As neural networks are designed to emulate the complex processes of biological neural systems, including learning, knowledge formation, and memory, statistical tuning provides a framework for systematically analyzing and improving these models. This process involves utilizing statistical methods to evaluate the models behavior, focusing on its training processes, parameters, and unit performance. By employing statistical tests to assess the interaction between the model and its environment such as how it responds to various inputs and feedback- statistical tuning helps in refining the models accuracy and effectiveness. This approach not only facilitates a better understanding of the models cognitive behavior but also ensures that the performance of ANNs aligns with predefined objectives. The methodologies discussed in the article, including accuracy bootstrapping, hidden neuron analysis, and dimension reduction techniques, are integral to this process, enabling a comprehensive assessment of model efficiency and performance.

**Definition 1** (ANNs Statistical Tuning) *Statistical tuning for ANNs is the process of using statistical methods to monitor and analyze the behavior of the neural network unite. This includes employing statistical model to represent the models behavior through training processes, model parameters, and unit performance. It also encompasses interaction with the surrounding environment, such as evaluating how the model responds to inputs and feedback. Based on the results of these statistical tests, the model is updated and improved to ensure its performance aligns with the desired objectives, thereby enhancing the understanding and effectiveness of neural network models.*

In this article, we will present methodologies related to , statistical tools of the importance of the model input, confidence intervals (CI) for ANN accuracy, hidden neuron number reduction, and hidden neuron analysis, focusing specifically on ANNs with a single hidden layer. The key contributions of this article are as follows:

The key contributions of this work are:

1. Developing an accuracy bootstrapping method for evaluating the performance of ANNs.

2. Analyzing hidden neurons through:

3. Statistical tests to evaluate the efficiency of neurons and group similar ones.

4. Applying logistic regression to assess the effectiveness of hidden neurons.

5. Testing the behavior of each hidden neuron $HN_i$ with respect to each output neuron and analyzing its performance.

6. Reducing the number of neurons in the neural network using clustering algorithms and *PCA*.

## 2   Important Definitions

To begin, it is essential to present a mathematical definition for the Feedforward Artificial Neural Network (FANN), as this is crucial for understanding and improving our analysis. This definition, introduced by [23] and [3], is given as follows:

**Definition 2** (Feedforward Artificial Neural Network) *Let $u_o$ denote the o-th input feature, $v_{ij}$ represent the weight connecting the j-th neuron in the previous layer to the i-th neuron in the current layer, and $g_l$ be the activation function of the l-th hidden layer. The output $z_k$ of a neural network with $m$ layers can be expressed as:*

$$z_k = g_m \left( \sum_l \left( v_{kl} g_{m-1} \left( ... g_2 \left( \sum_n \left( v_{rn} g_1 \left( \sum_o (v_{no} u_o) \right) \right) \right) \right) \right) \right) \tag{1}$$

*where, $z_k$ represents the k-th output of the network. The weight $v_{ij}$ denotes the connection from the j-th neuron in one layer to the i-th neuron in the subsequent layer, and $u_o$ signifies the o-th input feature. The activation functions are $g_1$ for the first hidden layer, $g_2$ for the second hidden layer, and $g_l$ for the l-th hidden layer. Finally, $g_m$ is the activation function for the m-th (output) layer.*

**Definition 3** (FANNs as a Nonlinear Regression Function) *[ 9 ] Consider a nonlinear regression function $f_0$ modeled*

by a fully-connected, single-layer feed-forward neural network $f$. This network is characterized by a bounded activation function $\psi$ on $\mathbb{R}$ and a set of $K$ hidden units. The function of the neural network can be expressed as:

$$f_0(x) = b_0 + \sum_{k=1}^{K} b_k \psi\left(a_{0,k} + a_k^\top x\right) \tag{2}$$

where, $b_0, b_k$, and $a_{0,k}$ are real-valued parameters, and $a_k \in \mathbb{R}^d$ represents the weight vector associated with the $k$-th hidden unit. Functions of this form are dense in $C(X)$, indicating that they are universal approximators. By selecting a sufficiently large dimension $K$ for the network, the function $f$ can approximate the target function $f_0$ to any desired level of precision.

## 3    Statistical Tuning of ANN

### 3.1    Reducing the Number of Hidden Neurons

ANNs often include numerous neurons within their hidden layers to capture complex patterns and relationships in data. However, not all neurons contribute equally to the models overall performance. Some neurons may produce redundant or noisy outputs, potentially decreasing the models efficiency by introducing unnecessary complexity. To address this challenge, we propose methods that aim to streamline the network without eliminating neurons entirely. One effective approach is to group neurons into clusters based on their outputs. By aggregating the outputs within each cluster and creating simplified representations, our goal is to maintain or even enhance the models performance while reducing its complexity. This clustering technique ensures that no neurons are discarded, thereby preserving the robustness and effectiveness of the network. Additionally, *PCA* provides an alternative method for reducing the number of neurons in a hidden layer. PCA focuses on identifying a smaller set of principal components that encapsulate the majority of the variance present in the neuron outputs. By retaining these principal components and discarding less significant ones, the complexity of the model can be reduced while striving to maintain or improve its performance. The following sections delve into these methods in detail. We first describe the clustering methods, which involve grouping neuron outputs into clusters to simplify the model while preserving its functionality. Next, we explore PCA techniques for reducing neuron counts by focusing on the principal components of neuron outputs. Both methods aim to enhance the efficiency of ANNs by optimizing the number of neurons in the hidden layers clustering methods do not remove or omit neurons; rather, they group the neurons into clusters. In this approach, we select a subset of neurons that we believe perform well and are sufficient for the models objectives.

### 3.1.1    Clustering methods

*Definition 4 Let $f: \mathbb{R}^n \to \mathbb{R}^p$ be a function represented by an ANN with one hidden layer containing $k$ neurons. Let $\boldsymbol{X} = [X_1, X_2, \ldots, X_k]$ is the outputs of the hidden layer, where each $X_i$ represents the output of the $i$-th neuron in the hidden layer. Then $\boldsymbol{X}$ can be clustered into $m$ groups, where $m < k$, such that each cluster $C_j$ represents an aggregated feature derived from the outputs within that cluster. Formally, the clustering results in new outputs $\chi = [\chi_1, \chi_2, \ldots, \chi_m]$, where each $\chi_j$ is a function of the outputs belonging to the $j$-th cluster. Then there exists at least one cluster output $\chi_i$ (or the set $\chi$ as a whole) such that the modes performance with the reduced representation $\chi$ is superior to or comparable with the performance of the original model using $\boldsymbol{X}$.*

—

The methodology involves two main steps: clustering of neuron outputs and construction of the reduced model. First, the neuron outputs $\{X_1, X_2, \ldots, X_k\}$ are represented as vectors in a high-dimensional space. A clustering algorithm, such as k-means, hierarchical clustering, or another suitable method, is then applied to group these outputs into m clusters, where m < k. For each cluster $C_j$, a new output $Y_j$ is created as an aggregated representation of the neurons within that cluster. This aggregation could be the centroid, mean, sum, or maximum of the outputs in the cluster. Next, the network architecture is adjusted by replacing the original neuron outputs $\boldsymbol{X}$ with the new aggregated outputs $\chi$. This modification changes the network to use m neurons instead of k, thereby reducing the complexity of the hidden layer. The reduced model is then trained, and its performance is compared with that of the original model on validation and test datasets to assess whether the reduced representation maintains or improves the models performance. For more details see algorithm 3.

### 3.1.2    PCA methods

This method offers an alternative approach to reducing the number of neurons in a hidden layer by employing *PCA* on the neuron outputs. The primary objective is to identify a smaller set of principal components that encapsulate most of the variance in the neuron outputs. By focusing on these principal components, the models complexity can be reduced while aiming to maintain or even enhance its performance. In this approach, we start by representing the outputs $\{X_1, X_2, ..., X_k\}$ as vectors in a high-dimensional space. *PCA* is then applied to these vectors to reduce their dimensionality. The outcome of this process is a set of principal components $\{X_1^*, X_2^*, ..., X_m^*\}$, where $m < k$. These principal components capture the majority of the variance present in the original outputs. To construct the reduced model, the original neuron outputs $\mathbf{X}$ are replaced with the new principal components $\mathbf{X}^*$ in the network architecture. This adjustment results in a model with m neurons instead of k, thereby simplifying the hidden layer. Following the model adjustment, the reduced model is trained and its performance is compared to that of the original model using validation and test datasets. This comparison helps to ensure that the reduction in complexity does not adversely affect the models performance.

*Definition 5 Let $f : \mathbb{R}^n \to \mathbb{R}^p$ be a function represented by an ANN with one hidden layer containing k neurons. The output of the hidden layer can be expressed as a vector $X = [X_1, X_2, ..., X_k]$, where each $X_i$ represents the output of the i-th neuron. By applying PCA to the outputs $\{X_1, X_2, ..., X_k\}$, we can obtain a reduced set of principal components $\{X_1^*, X_2^*, ..., X_m^*\}$, where $m < k$. These principal components capture the most significant variance in the original neurons outputs. The models performance using the reduced set of principal components $X^*$ is superior to or comparable with the performance of the original model using the full set of neuron outputs $X$.*

For more details see algorithm 5.

## 3.2 Accuracy confidence Interval

In evaluating the performance of ANN models, it is crucial to estimate the *CI* for accuracy metrics to understand the reliability of the models performance. Confidence intervals provide a range within which we expect the true accuracy of the model to lie, with a specified level of confidence. Two widely used methodologies for constructing these *CIs* are based on the Central Limit Theorem (CLT) and Bootstrap techniques. The CLT-based method involves calculating the sample statistic, such as the mean accuracy across multiple bootstrap samples, and then using statistical theory to construct the *CI*. This approach requires determining the confidence level, computing the critical value, and calculating the margin of error. The resulting *CI* provides an estimate of where the true model accuracy is likely to fall based on the distribution of the sample statistics. In contrast, the Bootstrap-based method involves generating numerous bootstrap samples from the original dataset and training the ANN model on each sample. By computing the accuracy for each bootstrap sample, we create a distribution of accuracies from which we can derive *CIs*. This approach is particularly useful for capturing the variability in model performance due to different sample configurations and offers a practical, data-driven means of estimating *CIs*. Both methods are instrumental in assessing the robustness of ANN models by providing insights into the variability and reliability of the accuracy estimates. The CLT-based method leverages theoretical properties of distributions, while the Bootstrap method provides empirical estimates based on observed data variations. Utilizing these methods ensures a comprehensive evaluation of the models performance and helps in making well-informed decisions based on statistical evidence.

### 3.2.1 CI Based on CLT

To estimate the *CI* for a sample statistic, such as model accuracy across bootstrap samples, a systematic procedure is followed. This approach involves calculating the sample statistic, determining the desired confidence level, computing the critical value, and calculating the margin of error. The final step is to construct the *CI*, providing a range where the true parameter value is expected to lie with a specified confidence level. First, calculate the sample statistic, which in this case is the mean accuracy across the bootstrap samples. The mean accuracy is computed as: $CC = \frac{1}{n} \sum_{i=1}^{n} ACC_i$, where $ACC_i$ is the accuracy for the $i$-th bootstrap sample, and $n$ is the total number of bootstrap samples. The standard deviation is then calculated as: $s = \sqrt{\frac{1}{n-1} \sum_{i=1}^{n} (ACC_i - ACC)^2}$. Next, determine the confidence level, such as 95%, which represents the likelihood that the *CI* will contain the true parameter value. The significance level $\alpha$ is computed as: $\alpha = 1$-ConfidenceLevel. Using this confidence level, we calculate the critical value, either from the Z-distribution (for large samples) or the t -distribution (for smaller samples). The critical Z -value is given as: $Z_{\text{critical}} = Z - \text{value at}$ $\left(1 - \frac{\alpha}{2}\right)$. The Z-value is taken from the Z-table based on the significance level $\alpha$. The margin of error is then calculated as: $\text{Margin of Error} = Z_{\text{critical}} \times \frac{s}{\sqrt{n}}$. Finally, the *CI* is constructed using the margin of error. The lower limit of the *CI* (CLI) is: $CLI = ACC - \text{Margin of Error}$ and the upper limit of the confidence interval (CUI) is: $CUI = ACC + \text{Margin of Error}$ Thus, the confidence interval $[CLI, CUI]$ provides the range where the true accuracy is likely to fall with the specified confidence level. For more details see algorithm [6].

### 3.2.2  CI Based on Bootstrap

To compute Bootstrap Confidence Intervals (CI) for model accuracy, the process begins by following one start with steps consist of training a base model and generating bootstrap model Based on samples. The following is an outline of this methodology: For each bootstrap sample, the model is trained using transfer learning, starting with pre-trained weights. The accuracy of the model for each bootstrap sample is then calculated. This accuracy is denoted by $ACC^{(i)}$ and computed as the proportion of correctly predicted labels out of the total number of samples. The mean accuracy is computed by averaging the accuracies obtained from all bootstrap samples. This serves as the central estimator for the models performance. Next, the differences between each bootstrap accuracy and the overall mean are calculated. These differences are sorted to prepare for percentile calculations. Lower and upper percentiles of the sorted differences are computed. These percentiles represent the bounds for calculating the $CI$, with the lower percentile corresponding to $100 \cdot \frac{\alpha}{2}\%$ and the upper percentile to $100 \cdot \left(1 - \frac{\alpha}{2}\right)\%$. Finally, the lower and upper bounds of the $CI$ are determined by adjusting the bootstrap mean accuracy using the percentiles. The resulting CI provides the range within which the true model accuracy is likely to fall with a given confidence level. This method thus constructs a statistical range that estimates the likely accuracy of the model while accounting for uncertainty in the sample data. For more details see algorithm 7.

## 3.3  Importance of the Neurons

To evaluate the importance of neurons, it is essential to consider their types, such as input neurons or hidden neurons. This section discusses statistical approaches for assessing the significance of these different types of neurons.

### 3.3.1  Importance of the Input Neurons

This problem has already been addressed by [9], whose findings are summarized by the algorithms 2.

### 3.3.2  Importance of Hidden Neurons

For each hidden neuron output $(X_i)_{i=1}^k$, we will analyze its output in relation to multiple outputs $(Y_i)_{i=1}^p$. This approach involves studying the contribution of each neuron $X_i$ to every output $Y_i$. To evaluate the importance of a neuron $X_i$ across multiple output nodes $\{Y_1, Y_2, \dots, Y_q\}$, begin by collecting the output values of $X_i$ for each output node. So, we perform a statistical test to compare these outputs, using methods such as ANOVA, t-tests, or other relevant statistical techniques. Evaluate the test statistic by comparing it against a chosen significance level $\alpha$. For more details see algorithm 2.

# 4  Application

## 4.1  Iris Model

The Iris dataset is seminal in machine learning and statistics, frequently utilized to illustrate data analysis techniques and algorithms. Introduced by Sir Ronald A. Fisher in 1936 [8], this dataset formed part of his research on discriminant analysis. It has since become a benchmark in the machine learning community for evaluating classification algorithms. The Iris dataset comprises 150 samples of iris flowers, each characterized by four features: Sepal Length (in centimeters), Sepal Width (in centimeters), Petal Length (in centimeters), and Petal Width (in centimeters). Each flower is classified into one of three species: Iris-setosa, Iris-versicolor, or Iris-virginica.

Table 1: ANN Model: IrisModel

| Layer (type) | Output Shape | # Params |
|---|---|---|
| input_layer  (InputLayer) | (None, 4) | 0 |
| hidden_layer  (Dense) | (None, 10) | 50 |
| Output_layer | (None, 3) | 33 |

The "ANN" model features three layers. The first layer, input layer, accepts input data with a shape of (None, 4), indicating that the network expects input with 4 features, with None allowing for variable batch sizes. This layer contains no trainable parameters. The second layer, hidden layer, is a dense (fully connected) layer with 10 neurons, producing an output shape of (None, 10) and including 50 trainable parameters. The parameter count is given by (inputfeatures $\times$ neurons) + neurons, which calculates to $(4 \times 10) + 10 = 50$. The third layer, output layer, is another dense layer with 3 neurons corresponding to the output classes. It produces an output shape of (None, 3) and contains 33 trainable parameters, calculated as $(10 \times 3) + 3 = 33$, where 10 represents the number of inputs from the previous layer, and 3 is the number of neurons. In total, the model has 83 trainable parameters, with an additional

168 parameters used by the optimizer for weight adjustment during training. There are no non-trainable parameters. The model is compact, with a total of 251 parameters, making it efficient for training and suitable for applications requiring a simple neural network architecture. The model achieved an accuracy of 96%, the outputs are detailed in Table 2. Figure 1 depicts the empirical probability density function and scatter plot for the outputs of the hidden layer.

Table 2: Hidden Neurons output of the Iris Test Dataset

| $N_1$ | $N_2$ | $N_3$ | $N_4$ | $N_5$ | $N_6$ | $N_7$ | $N_8$ | $N_9$ | $N_{10}$ |
|---|---|---|---|---|---|---|---|---|---|
| 0.124355 | 0.154244 | 0.460738 | 1.600283 | 0.004579 | 0.000000 | 0.000000 | 1.344940 | 0.000000 | 1.147511 |
| 1.707796 | 0.000000 | 2.874927 | 0.000000 | 0.000000 | 0.992714 | 0.750437 | 0.000000 | 3.287311 | 0.175460 |
| 0.000000 | 2.503711 | 0.000000 | 4.093118 | 1.976380 | 0.000000 | 0.000000 | 2.430906 | 0.000000 | 1.141091 |
| 0.000000 | 0.428091 | 0.212120 | 1.288688 | 0.441609 | 0.000000 | 0.000000 | 1.005065 | 0.000000 | 0.655154 |
| 0.000000 | 0.789750 | 0.591430 | 2.230830 | 0.434040 | 0.000000 | 0.000000 | 1.420828 | 0.000000 | 1.215945 |
| 2.025455 | 0.000000 | 1.939294 | 0.000000 | 0.000000 | 0.889137 | 0.948322 | 0.000000 | 3.017484 | 0.522979 |
| 0.501715 | 0.000000 | 0.215990 | 0.645844 | 0.059886 | 0.000000 | 0.069999 | 0.606518 | 0.095047 | 0.622777 |
| 0.000000 | 1.786332 | 0.078372 | 1.926190 | 1.822941 | 0.000000 | 0.000000 | 0.873577 | 0.000000 | 0.000000 |
| 0.216982 | 1.050481 | 0.000000 | 2.725117 | 0.474215 | 0.000000 | 0.000000 | 1.892343 | 0.000000 | 1.731530 |
| 0.575586 | 0.102159 | 0.118850 | 1.267909 | 0.000000 | 0.000000 | 0.000000 | 1.055809 | 0.000000 | 1.108014 |
| 0.000000 | 1.104305 | 0.388901 | 1.391543 | 1.296182 | 0.000000 | 0.000000 | 0.802474 | 0.000000 | 0.000000 |
| 2.860799 | 0.000000 | 1.323429 | 0.000000 | 0.000000 | 1.033653 | 1.524763 | 0.000000 | 3.261098 | 1.162466 |
| 2.228740 | 0.000000 | 2.386521 | 0.000000 | 0.000000 | 1.035738 | 0.993456 | 0.000000 | 3.376636 | 0.593891 |
| 2.726689 | 0.000000 | 1.542134 | 0.000000 | 0.000000 | 1.032774 | 1.438181 | 0.000000 | 3.256508 | 1.064779 |
| 1.972131 | 0.000000 | 2.556948 | 0.000000 | 0.000000 | 1.175545 | 1.230334 | 0.000000 | 3.862498 | 0.000000 |
| 0.000000 | 0.418078 | 0.911089 | 0.886180 | 0.650252 | 0.000000 | 0.000000 | 0.614369 | 0.000000 | 0.111514 |
| 0.000000 | 1.480567 | 0.000000 | 1.997023 | 1.604259 | 0.000000 | 0.000000 | 1.294941 | 0.000000 | 0.646789 |
| 2.235417 | 0.000000 | 2.179166 | 0.000000 | 0.000000 | 1.301748 | 0.769336 | 0.000000 | 2.579879 | 1.001058 |

Meanwhile, Figure 2 presents a heatmap of the correlation matrix.

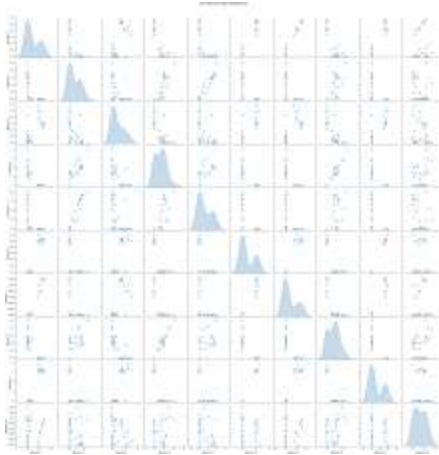

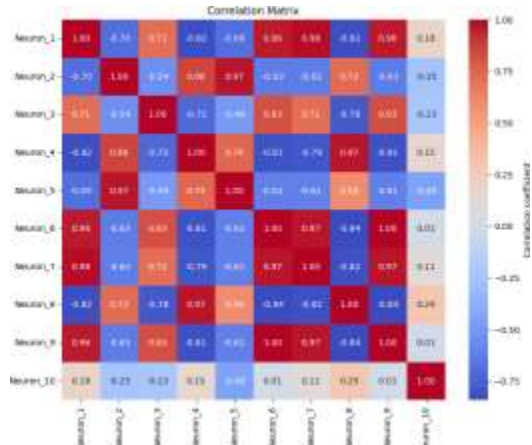

Figure 1: Empirical probability density function of the hidden layer outputs for Iris Dataset Model

Figure 2: Correlation matrix of the hidden layer outputs for Iris Dataset Model

### 4.1.1 Impact of Hidden Neurons on Output Neurons

To evaluate the significance of the impact of each hidden neuron $HN_k$ on the output neurons $Y_i$, the Mann-Whitney U Test was applied. The hypotheses are:

- **Null Hypothesis** ($H_0$): The distributions of the outputs from hidden neuron $HN_k$ to each output neuron $Y_i$ are identical.

- **Alternative Hypothesis** ($H_1$): There are significant differences in the distributions among at least two output neurons.

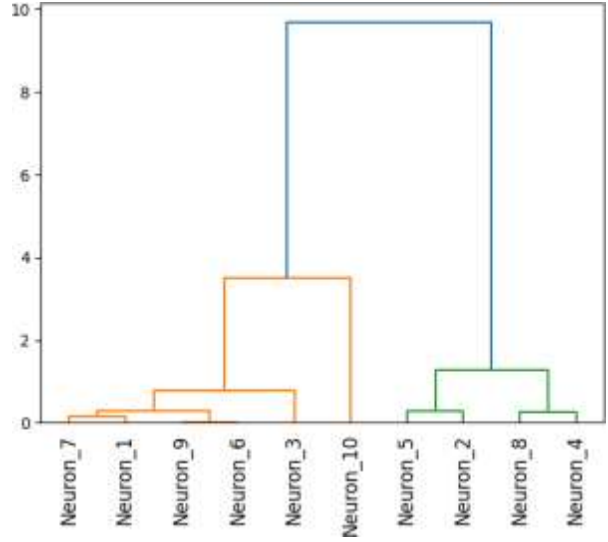

Figure 3: Hidden Layer Outputs Hierarchical Clustering

Table 3: Mann-Whitney U Test Results for Each Hidden Neuron to Output Neurons for Iris Dataset Model

| k | $O_{k1}$ vs $O_{k2}$ | | $O_{k1}$ vs $O_{k3}$ | | $O_{k2}$ vs $O_{k3}$ | |
|---|---|---|---|---|---|---|
| | U statistic | P-value | U statistic | P-value | U statistic | P-value |
| 0 | 316.0 | 0.0443 | 50.0 | 1.76e-09 | 50.0 | 1.76e-09 |
| 1 | 408.5 | 0.4828 | 719.5 | 4.14e-06 | 719.5 | 4.14e-06 |
| 2 | 719.5 | 4.14e-06 | 719.5 | 4.14e-06 | 408.5 | 0.4828 |
| 3 | 719.5 | 4.14e-06 | 719.5 | 4.14e-06 | 399.5 | 0.3921 |
| 4 | 300.0 | 0.0243 | 50.0 | 1.76e-09 | 50.0 | 1.76e-09 |
| 5 | 18.0 | 1.50e-10 | 277.0 | 0.0105 | 882.0 | 1.50e-10 |
| 6 | 308.0 | 0.0306 | 334.0 | 0.0776 | 559.0 | 0.0973 |
| 7 | 738.0 | 1.59e-06 | 418.0 | 0.5990 | 162.0 | 1.59e-06 |
| 8 | 481.0 | 0.6244 | 772.0 | 2.46e-07 | 772.0 | 2.46e-07 |
| 9 | 0.0 | 3.02e-11 | 900.0 | 3.02e-11 | 900.0 | 3.02e-11 |

Table 3 summarizes the results of the Mann-Whitney U test performed to evaluate the differences between hidden neurons and output neurons. The test compares the distributions of output neurons for each hidden neuron, providing a U statistic and a p-value for each pairwise comparison. Several key observations arise from the results:

- **Hidden neuron 0:** All comparisons between output neurons $O_{k1}$ vs $O_{k2}$, $O_{k1}$ vs $O_{k3}$, and $O_{k2}$ vs $O_{k3}$ yield significant differences, with p-values below the standard threshold of 0.05.

- **Hidden neuron 5**: Exhibits highly significant differences across all comparisons, with p-values far lower than 0.05, indicating strong evidence of differences in output distributions.

- **Hidden neuron 9**: Displays significant results in all comparisons, with extremely low p-values, showing clear distinctions between the output neuron distributions.

In contrast:

- **Hidden neuron 1**: The p-value for the comparison $O_{k1}$ vs $O_{k2}$ is 0.4828, which is well above 0.05, indicating no significant difference between these output neurons. Similar patterns are seen in other comparisons for neuron 1, suggesting this neuron does not effectively distinguish between output classes.

- **Hidden neurons 6 and 8**: Show mixed results, with some comparisons yielding p-values greater than 0.05, as highlighted by the gray cells in the table, indicating inconsistent differentiation between output classes.

The results of the Mann-Whitney U test show that hidden neurons 0, 5, and 9 significantly distinguish between the output neurons' distributions, whereas neurons 1, 6, and 8 exhibit less clear differentiation. So, neurons with p-values greater than 0.05 do not differentiate between output classes $O_{ki}$ and $O_{kj}$ For instance, hidden neuron $HN_1$ has a p-value of 0.4828 for the comparison between $O_{k1}$ and $O_{k2}$, indicating that it does not differentiate between classes $Y_1$ and $Y_2$. This suggests that the outputs of $HN_1$ are drawn from the same distribution, meaning this neuron output weights do not show significant variation between these two classes.

### 4.1.2 Clustering of Hidden Layer Neurons

Figure 3 illustrates the clustering of the hidden layer neurons into three distinct clusters. The details of these clustering results are summarized as follows: The neurons can be grouped into clusters as sets:

- **Cluster 1**: {Neuron 2, Neuron 4, Neuron 5, Neuron 8}

- **Cluster 2**: {Neuron 1, Neuron 3, Neuron 6, Neuron 7, Neuron 9}

- **Cluster 3**: {Neuron 10}

The **Cluster 1 Model** features two dense layers: the first with 4 output values and 20 parameters, and the second with 3 output values and 15 parameters. This model has a total of 107 parameters, including 35 trainable parameters and 432 bytes of memory. The **Cluster 2 Model** includes two dense layers: one with 5 output values and 25 parameters, and another with 3 output values and 18 parameters. This model totals 131 parameters, with 43 trainable parameters and 528 bytes of memory. The **Cluster 3 Model** comprises two dense layers: the first with 1 output value and 5 parameters, and the second with 3 output values and 6 parameters. The model has 35 parameters, with 11 trainable parameters and 144 bytes of memory. In terms of accuracy, **Cluster 1** achieved a perfect accuracy of 1.00, **Cluster 2** attained an accuracy of 0.97, and **Cluster 3** recorded an accuracy of 0.73.

### 4.1.3 95% Confidence Intervals

**CLT Method**: The mean accuracy, denoted as $\overline{ACC}$, is 0.9614 with a standard deviation of 0.0352. Using a critical t-value of 1.9623, the *CI* is computed as [0.9592, 0.9636].
**Bootstrap Method**: The *CI* is estimated as [0.87, 0.97].

## 4.2 IDC Dataset

The IDC dataset [12] consists of digitized histopathology slides of breast cancer (BCA) from 162 women diagnosed with invasive ductal carcinoma (IDC) at the Hospital of the University of Pennsylvania and The Cancer Institute of New Jersey. These slides were captured using a whole-slide scanner at 40x magnification, achieving a resolution of 0.25 µm/pixel. Given the enormous size of these whole-slide images, which can exceed $10^{10}$ pixels, direct analysis is impractical. Therefore, each whole-slide image (WSI) was down sampled by a factor of 16:1, resulting in a resolution of 4 µm/pixel. The IDC dataset contains two labels with the following counts: IDC(-) has 198,738 instances, while IDC(+) has 78,786 instances. The dataset is divided into two subsets: 80% for training the models and 20% for testing and evaluating model performance.

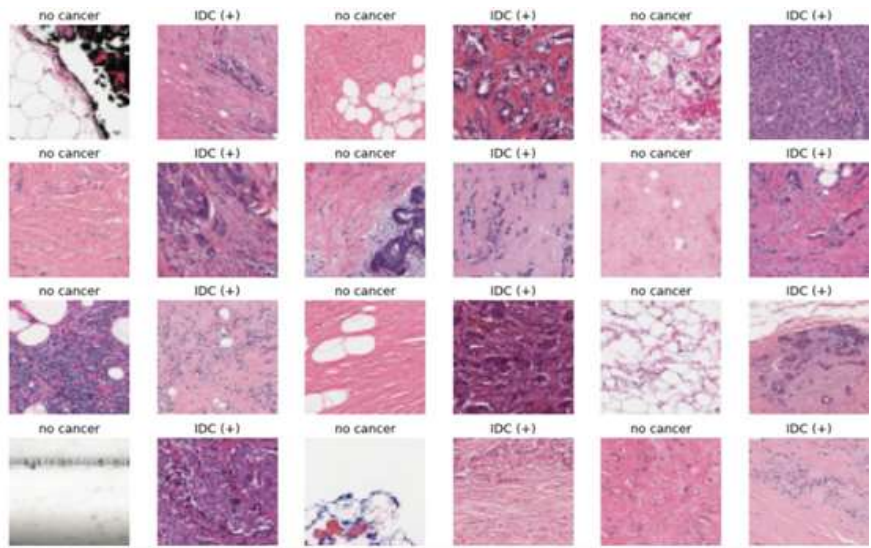

Figure 4: IDC Dataset

#### 4.2.1 Deep Convolutional Models

Deep convolutional models are employed to extract features from the images, transforming the three-dimensional pixel matrix into a numerical vector in $R^n$. The architecture of our model starts with an input layer that accepts images of size 50x50 pixels with 3 color channels, typically representing an RGB image. The first convolutional block consists of a convolutional layer applying 32 filters of size 3x3, followed by the ReLU activation function. This is followed by a batch normalization layer. Another convolutional layer applies 32 filters of size 3x3 with ReLU activation. The output is processed through a max pooling layer with a 2x2 filter, reducing the spatial dimensions, and another batch normalization layer. Subsequently, the network includes dropout for regularization, randomly dropping some units. The second convolutional block begins with a convolutional layer applying 64 filters of size 3x3 with ReLU activation, followed by batch normalization. Another convolutional layer applies 64 filters of size 3x3 with ReLU activation, followed by a batch normalization layer. This block concludes with a second max pooling layer to further reduce the spatial dimensions. The third convolutional block starts with another dropout layer for additional regularization. It then applies a convolutional layer with 128 filters of size 3x3 using ReLU activation, followed by batch normalization. After the convolutional blocks, the output is flattened into a one-dimensional vector and passed through a fully connected layer with 128 units and ReLU activation, followed by batch normalization. The architecture ends with the output layers. A dropout layer is applied to prevent overfitting, followed by a dense layer with 64 units using ReLU activation. Another dense layer with 24 units follows, also with ReLU activation. The model concludes with a final dense layer containing 2 units with softmax activation, indicating that the network is solving a classification problem with 2 classes. The results of this model are shown in Table 8.

Table 4: Classification report of the proposed CNN model for the IDC dataset.

| Index | Precision | Recall | F1-Score | BAC | Sample Size |
|---|---|---|---|---|---|
| Negative | 0.9351 | 0.9021 | 0.9183 | 0.9186 | 39,719 |
| Positive | 0.7739 | 0.8426 | 0.8068 | 0.8082 | 15,786 |
| Accuracy | 0.8852 | 0.8852 | 0.8852 | 0.8852 | - |
| Macro Avg | 0.8545 | 0.8724 | 0.8626 | 0.8634 | 55,505 |
| Weighted Avg | 0.8893 | 0.8852 | 0.8866 | 0.8872 | 55,505 |

#### 4.2.2 Logistic Regression

To apply logistic regression models, we use the outputs from the Top layer in the denes of the CNN model. Each image is represented by 24 independent variables. The descriptive statistics for these variables are shown in Table 5. The design matrix $X$ is defined as: $X = (\mathbb{I}_n, X_1, X_2, \ldots, X_{24})$, where $\mathbb{I}_n = (1,1,\ldots,1)^T$ and $X_i = (x_1^{(i)}, x_2^{(i)}, \ldots, x_n^{(i)})$ represents the output of the $i$-th neuron. Figure [4] illustrates the correlation of the 24 neuron outputs. Figure 4 illustrates the correlation of the 24 neuron outputs.

Table 5: Descriptive statistics of the 24 independent variables extracted by the proposed CNN for the IDC dataset.

| Variable | Mean | Std Dev | Q1 | Median | Q3 | Max |
|---|---|---|---|---|---|---|
| $X_1$ | 0.006979 | 0.048053 | 0 | 0 | 0 | 1.450578 |
| $X_2$ | 0.621080 | 1.159904 | 0 | 0 | 0.818471 | 7.492388 |
| $X_3$ | 0.975385 | 1.473224 | 0 | 0 | 1.662232 | 26.60136 |
| $X_4$ | 3.448834 | 4.149483 | 0 | 1.794995 | 5.903279 | 202.0401 |
| $X_5$ | 0.683227 | 0.992003 | 0 | 0.131018 | 1.126925 | 57.70039 |
| $X_6$ | 0.050156 | 0.178944 | 0 | 0 | 0 | 30.72196 |
| $X_7$ | 3.300140 | 2.551751 | 1.523839 | 2.700684 | 4.452011 | 190.4687 |
| $X_8$ | 0.040165 | 0.125586 | 0 | 0 | 0 | 1.583805 |
| $X_9$ | 1.805387 | 2.248423 | 0 | 0.817885 | 3.106404 | 23.47421 |
| $X_{10}$ | 2.141062 | 3.172278 | 0 | 0.611011 | 3.402364 | 241.2344 |
| $X_{11}$ | 0.317287 | 0.660116 | 0 | 0 | 0.175377 | 4.675481 |
| $X_{12}$ | 0.002884 | 0.029940 | 0 | 0 | 0 | 1.057262 |
| $X_{13}$ | 0.163499 | 0.374551 | 0 | 0 | 0.079030 | 5.224789 |
| $X_{14}$ | 1.353993 | 2.500610 | 0 | 0 | 1.719606 | 14.44952 |
| $X_{15}$ | 0.127713 | 0.317607 | 0 | 0 | 0 | 3.428244 |
| $X_{16}$ | 1.929913 | 1.439368 | 0.848341 | 1.627364 | 2.760290 | 22.44036 |
| $X_{17}$ | 0.246951 | 0.567942 | 0 | 0 | 0.102887 | 7.786866 |
| $X_{18}$ | 0.146410 | 0.331418 | 0 | 0 | 0.092526 | 2.764073 |
| $X_{19}$ | 0.243165 | 0.417401 | 0 | 0 | 0.156283 | 2.698198 |
| $X_{20}$ | 0.061496 | 0.147388 | 0 | 0 | 0.027969 | 1.887274 |

| | | | | | |
|---|---|---|---|---|---|
| $X_{21}$ | 0.106918 | 0.305371 | 0 | 0 | 0 | 4.022273 |
| $X_{22}$ | 0.356947 | 0.829875 | 0 | 0 | 0 | 8.743048 |
| $X_{23}$ | 0.189712 | 0.382960 | 0 | 0 | 0 | 3.647056 |
| $X_{24}$ | 0.189274 | 0.511094 | 0 | 0 | 0.043147 | 6.636074 |

Table 6: Logistic Regression Model Evaluation for the IDC dataset.

| Pseudo R-squared | AIC | BIC | No. Observations | Log-Likelihood |
|---|---|---|---|---|
| 0.772 | 60332.6969 | 60580.1494 | 222019 | -30142 |

| Df Model | LL-Null | Df Residuals | LLR p-value | |
|---|---|---|---|---|
| 23 | -1.32E+05 | 221995 | 0 | |

Table 7: Maximum likelihood estimators (MLE), p-values, and 95% $CIs$ for the coefficients of the logistic regression model for the IDC dataset.

| $\beta_i$ | Coef. | Std. Err. | z | P-value | [0.025, | 0.975] |
|---|---|---|---|---|---|---|
| $\beta_0$ | -0.4315 | 0.0259 | -16.68 | 1.69E-62 | -0.4800 | -0.3808 |
| $\beta_1$ | -0.1123 | 0.2806 | -0.40 | 0.68904 | -0.6600 | 0.4376 |
| $\beta_2$ | 0.6523 | 0.0336 | 19.39 | 9.16E-84 | 0.5860 | 0.7183 |
| $\beta_3$ | -0.3351 | 0.0496 | -6.76 | 1.42E-11 | -0.4300 | -0.2379 |
| $\beta_4$ | -0.3950 | 0.0408 | -9.69 | 3.22E-22 | -0.4700 | -0.3151 |
| $\beta_5$ | 0.0323 | 0.0503 | 0.64 | 0.52074 | -0.0700 | 0.1310 |
| $\beta_6$ | 0.2281 | 0.1351 | 1.69 | 0.09126 | -0.0400 | 0.4928 |
| $\beta_7$ | -0.5517 | 0.0250 | -22.02 | 0.00000 | -0.6000 | -0.5026 |
| $\beta_8$ | 0.1756 | 0.0940 | 1.87 | 0.06193 | -0.0100 | 0.3599 |
| $\beta_9$ | -0.2767 | 0.0377 | -7.33 | 2.25E-13 | -0.3500 | -0.2028 |
| $\beta_{10}$ | 0.4680 | 0.0449 | 10.42 | 1.94E-25 | 0.3800 | 0.5560 |
| $\beta_{11}$ | 0.3277 | 0.0504 | 6.50 | 7.98E-11 | 0.2300 | 0.4265 |
| $\beta_{12}$ | -1.4060 | 0.4462 | -3.15 | 0.00163 | -2.2800 | -0.5315 |
| $\beta_{13}$ | -0.6442 | 0.1161 | -5.55 | 2.87E-08 | -0.8700 | -0.4167 |
| $\beta_{14}$ | 0.2922 | 0.0235 | 12.46 | 1.30E-35 | 0.2460 | 0.3382 |
| $\beta_{15}$ | 1.2223 | 0.0587 | 20.83 | 2.31E-96 | 1.1070 | 1.3373 |
| $\beta_{16}$ | -0.1984 | 0.0398 | -4.98 | 6.41E-07 | -0.2800 | -0.1203 |
| $\beta_{17}$ | 0.7422 | 0.0381 | 19.46 | 2.50E-84 | 0.6670 | 0.8170 |
| $\beta_{18}$ | 0.1116 | 0.0490 | 2.28 | 0.02267 | 0.0160 | 0.2076 |
| $\beta_{19}$ | -0.6631 | 0.0485 | -13.67 | 1.44E-42 | -0.7600 | -0.5681 |
| $\beta_{20}$ | -0.1028 | 0.0505 | -2.04 | 0.04172 | -0.2000 | -0.0039 |
| $\beta_{21}$ | -0.4424 | 0.0470 | -9.42 | 4.47E-21 | -0.5300 | -0.3503 |
| $\beta_{22}$ | 1.7764 | 0.0939 | 18.92 | 8.31E-80 | 1.5920 | 1.9604 |
| $\beta_{23}$ | -0.1330 | 0.0471 | -2.82 | 0.00478 | -0.2300 | -0.0406 |
| $\beta_{24}$ | 0.0475 | 0.0651 | 0.73 | 0.46563 | -0.0800 | 0.1750 |

Table 8: Classification report of the proposed logistic regression model for the IDC dataset.

| | Precision | Recall | F1-Score | Sample Size |
|---|---|---|---|---|
| Negative | 0.921928 | 0.918679 | 0.920301 | 39719 |
| Positive | 0.797187 | 0.804257 | 0.800706 | 15786 |
| Accuracy | | | 0.886136 | |
| Macro avg | 0.859558 | 0.861468 | 0.860503 | 55505 |

| Weighted avg | 0.886451 | 0.886136 | 0.886287 | 55505 |

### 4.2.3   Logistic Regression Model Using Principal Components

*PCA* was applied to the 24 independent variables derived from the CNN model. Logistic regression models were then fitted for each combination of principal components. The results are summarized in Table 9, which includes the number of principal components used, the explained variance, and the accuracy of the logistic regression model.

Based on table 9, it is evident that the 6-component PCA model achieves the best accuracy. The performance of this model is detailed in the confusion matrix and classification report provided in Table 10.

## 5   Conclusion and Discussion

### 5.1   Comparison of Confidence Intervals

Based on the experimental study detailed in (4.1.2), the 95% CI based on the *CLT* is narrower ([0.9592, 0.9636]) compared to the bootstrap CI ([0.8700, 0.9700]). The narrower interval from the CLT suggests higher precision in estimating the mean accuracy, assuming that the data adheres to normality and sample size requirements. Conversely, the broader interval from the bootstrap method reflects greater variability and accommodates potential deviations from normality, thus providing a more flexible, data-driven estimate.

Table 9: Accuracy and Explained Variance for Different Number of Components for the IDC dataset.

| Components | Accuracy | Explained Variance | Components | Accuracy | Explained Variance |
|---|---|---|---|---|---|
| 1 | 0.88255 | 0.66433 | 13 | 0.88695 | 0.99598 |
| 2 | 0.88824 | 0.80139 | 14 | 0.88599 | 0.99705 |
| 3 | 0.88855 | 0.90599 | 15 | 0.88603 | 0.99805 |
| 4 | 0.88862 | 0.94761 | 16 | 0.88594 | 0.99856 |
| 5 | 0.88859 | 0.95759 | 17 | 0.88624 | 0.99901 |
| 6 | 0.88860 | 0.96599 | 18 | 0.88617 | 0.99935 |
| 7 | 0.88808 | 0.97316 | 19 | 0.88608 | 0.99963 |
| 8 | 0.88797 | 0.97928 | 20 | 0.88614 | 0.99969 |
| 9 | 0.88796 | 0.98459 | 21 | 0.88621 | 0.99985 |
| 10 | 0.88805 | 0.98853 | 22 | 0.88614 | 0.99997 |
| 11 | 0.88815 | 0.99189 | 23 | 0.88612 | 0.99999 |
| 12 | 0.88709 | 0.99418 | 24 | 0.88614 | 1.00000 |

Table 10: Classification report for the 6-component LR-PCA model for the IDC dataset.

| Metric | Precision | Recall | F1-score | Support |
|---|---|---|---|---|
| Negative | 0.9249 | 0.9189 | 0.9219 | 39,719 |
| Positive | 0.7993 | 0.8124 | 0.8058 | 15,786 |
| Accuracy | 0.8886 | 0.8886 | 0.8886 | - |
| Macro Avg | 0.8621 | 0.8656 | 0.8638 | 55,505 |
| Weighted Avg | 0.8892 | 0.8886 | 0.8889 | 55,505 |

The substantial difference between the intervals underscores the influence of the chosen method on CI width. The CLT-based CI presumes a well-behaved sample distribution, while the bootstrap method, not reliant on such assumptions, offers a potentially wider range that might better capture data variability. Therefore, while the CLT-based CI provides a precise estimate under ideal conditions, the bootstrap CI delivers a more robust estimate in the presence of data variability or non-normality. The selection of the method should align with the data characteristics and the assumptions that can be reasonably made.

### 5.2   Effect Measurement of Hidden Neurons on Output Neurons

Based on the experiment outlined in 3, the analysis using the Mann-Whitney U test reveals important insights into the contribution of hidden neurons to class distinction within the neural network. The results indicate that certain neurons, specifically $HN_0$, $HN_5$, and $HN_9$, exhibit significant differences in output distributions, demonstrating their effectiveness in differentiating between output classes. These neurons play a crucial role in capturing the relevant features needed for accurate classification.

In contrast, neurons such as $HN_1$, $HN_6$, and $HN_8$ show less consistent performance. For these neurons, the p-values for some comparisons are above the significance threshold, suggesting that they do not effectively distinguish between certain classes. This variability implies that while some neurons contribute meaningfully to the classification process, others may offer minimal benefit or introduce redundancy.

These findings underscore the importance of evaluating the role of individual neurons in complex models. By identifying neurons that do not significantly contribute to class separation, one can focus on refining or pruning these less impactful neurons. This approach can lead to a more efficient network design, enhancing the models overall performance and its ability to generalize and accurately classify unseen data.

### 5.3    Clustering Method

In the study detailed in 4.1.2, the comparison of parameter counts between models for different clusters and the original model reveals several key observations. For Cluster 1, the model has more parameters (107) compared to the original model (83), representing an increase of approximately 28.9%. Similarly, the Cluster 2 model, with 131 parameters, has about 57.8% more parameters than the original model, indicating greater complexity. In contrast, the Cluster 3 model has 35 parameters, which is 57.8% fewer than the original model, suggesting a simpler approach.

Overall, models for Clusters 1 and 2 are more complex than the original model, while the Cluster 3 model is simpler, highlighting different complexities and potential trade-offs between model simplicity and performance.

### 5.4    LR and PCA Approach

Based on the p-values provided in Table 7, the coefficients that lack statistical significance, as indicated by p-values exceeding 0.05, are $\beta_1$ (p-value = 0.68904), $\beta_5$ (p-value = 0.52074), $\beta_6$ (p-value = 0.09126), $\beta_8$ (p-value = 0.06193), and $\beta_{24}$ (p-value = 0.46563). These coefficients do not provide sufficient evidence to reject the null hypothesis at the standard significance level of 0.05.

On the other hand, based on the experimental study in 4.2.3, we compared our proposed models with those introduced by Reza and Ma (2018) and sampling methods of Janowczyk et al. (2016) as follows:

Table 11: Performance results and classification reports from various models for the IDC dataset.

| Method | Authors | F-score |
|---|---|---|
| 20x | Janowczyk et al. (2016) | 0.80 |
| + Dropout | Janowczyk et al. (2016) | 0.79 |
| ResNet34 | Janowczyk et al. (2016) | 0.79 |
| Imbalanced data | Reza (2018) | 0.7359 |
| Under-sampling | Reza (2018) | 0.8194 |
| Over-sampling (WR) | Reza (2018) | 0.8443 |
| ADASYN | Reza (2018) | 0.8402 |
| SMOTE | Reza (2018) | 0.8478 |
| The proposed CNN Model | - | 0.8852 |
| The proposed Logistic Regression | - | 0.8861 |
| The proposed 6 component PCA Model | - | 0.8886 |

The proposed models demonstrate nearly identical accuracy levels; however, our Logistic Regression (LR) and (LR-PCA) models are more interpretable, achieving the same accuracy while effectively meeting our objective for HN reduction.

# References

[1] Al-Mohamad, M. Y. (2024). *Artificial Neural Network Analysis Using Statistical Approaches*. MSc thesis, University of Tabriz.

[2] Alrashidi, M., Selamat, A., Ibrahim, R., & Krejcar, O. (2023). Social Recommendation for Social Networks Using Deep Learning Approach: A Systematic Review, Taxonomy, Issues, and Future Directions. IEEE Access.

[3] Angelov, P. P., Soares, E. A., Jiang, R., Arnold, N. I., & Atkinson, P. M. (2021). Explainable artificial intelligence: an analytical review. Wiley Interdisciplinary Reviews: Data Mining and Knowledge Discovery, 11(5), e1424.

[4] Carney, J. G., Cunningham, P., & Bhagwan, U. (1999, July). Confidence and prediction intervals for neural network ensembles. In *IJCNN'99. International Joint Conference on Neural Networks. Proceedings (Cat. No. 99CH36339)* (Vol. 2, pp. 1215-1218). IEEE.

[5] Chillotti, I., Joye, M., & Paillier, P. (2021). Programmable bootstrapping enables efficient homomorphic inference of deep neural networks. In *Cyber Security Cryptography and Machine Learning: 5th International Symposium, CSCML 2021, Be'er Sheva, Israel, July 89, 2021, Proceedings 5* (pp. 1-19). Springer International Publishing.

[6] Didona, D., & Romano, P. (2014). On bootstrapping machine learning performance predictors via analytical models. *arXiv preprint arXiv:1410.5102*.

[7] Doshi-Velez, F., & Kim, B. (2017). Towards a rigorous science of interpretable machine learning. arXiv preprint arXiv:1702.08608.

[8] Fisher, R. (1936). Iris. UCI Machine Learning Repository. Retrieved from https://doi.org/10.24432/C56C76.

[9] Horel, E., & Giesecke, K. (2020). Significance tests for neural networks. *Journal of Machine Learning Research*, 21(227), 1-29.

[10] Huang, C. G., Huang, H. Z., Li, Y. F., & Peng, W. (2021). A novel deep convolutional neural network-bootstrap integrated method for RUL prediction of rolling bearing. *Journal of Manufacturing Systems, 61*, 757-772.

[11] Janowczyk, A., & Madabhushi, A. (2016). Deep learning for digital pathology image analysis: A comprehensive tutorial with selected use cases. *Journal of Pathology Informatics*, 7(1), 29.

[12] Kasikrit. *IDC dataset*. Kaggle, 2021. Available at: https://www.kaggle.com/datasets/kasikrit/idc-dataset.

[13] Kononenko, I. (1989). Bayesian neural networks. Biological Cybernetics, 61(5), 361-370.

[14] Li, H., Tian, S., Li, Y., Fang, Q., Tan, R., Pan, Y., Huang, C., Xu, Y., & Gao, X. (2020). Modern deep learning in bioinformatics. Journal of molecular cell biology, 12(11), 823-827.

[15] Medeiros, M. C., Teräsvirta, T., & Rech, G. (2006). Building neural network models for time series: a statistical approach. Journal of Forecasting, 25(1), 49-75.

[16] Michelucci, U., & Venturini, F. (2021). Estimating neural networks performance with bootstrap: A tutorial. *Machine Learning and Knowledge Extraction, 3*(2), 357-373.

[17] Nixon, J., Tran, D., & Lakshminarayanan, B. (2020, December). Why arent bootstrapped neural networks better?. In *34th Conference on Neural Information Processing Systems (NeurIPS 2020)*, Virtual-only Conference.

[18] Reed, S., Lee, H., Anguelov, D., Szegedy, C., Erhan, D., & Rabinovich, A. (2014). Training deep neural networks on noisy labels with bootstrapping. *arXiv preprint arXiv:1412.6596*.

[19] Reza, M. S., & Ma, J. (2018, August). Imbalanced histopathological breast cancer image classification with convolutional neural network. In *2018 14th IEEE International Conference on Signal Processing (ICSP)* (pp. 619-624). IEEE.

[20] Reza, M. S., & Ma, J. (2018, August). Imbalanced histopathological breast cancer image classification with convolutional neural network. In *2018 14th IEEE International Conference on Signal Processing (ICSP)* (pp. 619-624). IEEE.

[21] Rudin, C. (2019). Stop explaining black box machine learning models for high stakes decisions and use interpretable models instead. Nature Machine Intelligence, 1(5), 206-215.

[22] Sharma, S. K., & Tiwari, K. N. (2009). Bootstrap based artificial neural network (BANN) analysis for hierarchical prediction of monthly runoff in Upper Damodar Valley Catchment. *Journal of Hydrology, 374*(3-4), 209-222.

[23] Samek, W., Montavon, G., Lapuschkin, S., Anders, C. J., & Müller, K.-R. (2021). Explaining deep neural networks and beyond: A review of methods and applications. Proceedings of the IEEE, 109(3), 247-278.

[24] Secchi, P., Zio, E., & Di Maio, F. (2008). Quantifying uncertainties in the estimation of safety parameters by using bootstrapped artificial neural networks. *Annals of Nuclear Energy, 35*(12), 2338-2350.

[25] Seoni, S., Jahmunah, V., Salvi, M., Barua, P. D., Molinari, F., & Acharya, U. R. (2023). Application of uncertainty quantification to artificial intelligence in healthcare: A review of last decade (20132023). Computers in Biology and Medicine, 107441.

[26] Soares, E., Angelov, P., Costa, B., & Castro, M. (2019). Actively semi-supervised deep rule-based classifier applied to adverse driving scenarios. 2019 International Joint Conference on Neural Networks (IJCNN).

[27] Specht, D. F. (1990). Probabilistic neural networks. Neural networks, 3(1), 109-118.

[28] Wang, Y.-X., & Hebert, M. (2016). Learning to learn: Model regression networks for easy small sample learning. Computer VisionECCV 2016: 14th European Conference, Amsterdam, The Netherlands, October 11-14, 2016, Proceedings, Part VI, 14.

[29] Weigend, A. S., & LeBaron, B. (1994, October). Evaluating neural network predictors by bootstrapping. In *Proc. Int. Conf. Neural Inform. Processing* (Vol. 2, pp. 1207-1212).

[30] Yu, Z., Wang, K., Wan, Z., Xie, S., & Lv, Z. (2023). Popular deep learning algorithms for disease prediction: a review. Cluster Computing, 26(2), 1231-1251.

[31] Zheng, Y., Xu, Z., & Xiao, A. (2023). Deep learning in economics: a systematic and critical review. Artificial Intelligence Review, 1-43.

# 6  Appendix

The PY code of application discussed can be accessed at the following GitHub repository:
https://github.com/yamenetoo/Artificial-Neural-Network-Analysis-Using-Statistical-Approaches

---

**Algorithm 1** Test Procedure for Hidden Neurons to output

---

1: **Input:** $HN\_outputs$, $O\_classes$, $\alpha$
2: **Output:** Results table with U statistics and p-values
3: Initialize empty results table $R$
4: **for** each hidden neuron $k$ in $HN\_outputs$ **do**
5:     Set $neuron\_outputs \leftarrow HN\_outputs[k]$
6:     **for** each pair of output classes $(O_{ki}, O_{kj})$ in $O\_classes$ **do**
7:         Extract $HN\_outputs[k][O_{ki}]$ and $HN\_outputs[k][O_{kj}]$
8:         Perform Mann-Whitney U test on $HN\_outputs[k][O_{ki}]$ and $HN\_outputs[k][O_{kj}]$
9:         Compute U statistic and p-value
10:        Store U statistic and p-value in $R[k]$ for the pair $(O_{ki}, O_{kj})$
11:     **end for**
12: **end for**
13: **for** each entry in $R$ **do**
14:     **if** $p\_value > \alpha$ **then**
15:         Mark the comparison as *not significant*
16:     **else**
17:         Mark the comparison as *significant*
18:     **end if**
19: **end for**
20: **Return** results table $R$

---

---

**Algorithm 2** Approximation of Limiting Distribution Using the Discretization Approach

---

1: **Input:** Function space $\Theta$, approximation parameter $\epsilon$, number of samples $m_N$, truncation order $N$
2: **Output:** Empirical approximate quantile function $F_{m_N,N}^{-1}(\alpha)$
3: **Step 1: Generate an $\epsilon$-Cover of $\Theta$**
4: Generate a finite set $\{f_1, f_2, \ldots, f_C\} \subset \Theta$ such that $\{f_1, f_2, \ldots, f_C\}$ is an $\epsilon$-cover of $\Theta$. ▷ This set approximates the function space $\Theta$ within $\epsilon$.
5: **Step 2: Compute Covariance Matrix**
6: **for** $i, j = 1$ to $C$ **do**
7:      Compute the covariance matrix element $\Sigma_{i,j} = \mathbb{E}_X[f_i(X)f_j(X)]$
8: **end for**
9: **Step 3: Sample from Multivariate Gaussian Distribution**
10: **for** $\omega = 1$ to $m_N$ **do**
11:      Sample $\mathbf{Z}_\omega = (Z_1^\omega, Z_2^\omega, \ldots, Z_C^\omega)$ from a multivariate Gaussian distribution $\mathcal{N}(0, \Sigma)$
12:      Compute the argmax $h_\omega^\star = \arg\max_{f_i \in \{f_1, \ldots, f_C\}} Z_i^\omega$      ▷ Find the function that maximizes the Gaussian process for the given sample.
13: **end for**
14: **Step 4: Approximate the Limiting Distribution**
15: **for** $\omega = 1$ to $m_N$ **do**
16:      Compute the test statistic $T_\omega = Apply functional(6) to h_\omega^\star$      ▷ Generate an approximate sample from the limiting distribution.
17: **end for**
18: **Step 5: Construct Empirical Distribution**
19: Sort the values $\{T_1, T_2, \ldots, T_{m_N}\}$ in ascending order
20: **for** $i = 1$ to $m_N$ **do**
21:      Define the empirical quantile function $F_{m_N,N}^{-1}(\alpha)$ such that $F_{m_N,N}^{-1}(\alpha) = T_i$ for $\alpha \in \left(\frac{i-1}{m_N}, \frac{i}{m_N}\right]$
22: **end for**
23: **Step 6: Evaluate Test Statistic**
24: Calculate the empirical test statistic $\hat{\lambda}_n$
25: **if** $\hat{\lambda}_n > F_{m_N,N}^{-1}(1 - \alpha)$ **then**
26:      Reject the null hypothesis
27: **else**
28:      Fail to reject the null hypothesis
29: **end if**
30: **Step 7: Control Accuracy with $\epsilon$**
31: Choose $\epsilon$ and $N$ such that the approximation error is within the desired level

---

$-$

---

**Algorithm 3** Neuron Clustering

---
1:  **Input:** Outputs of hidden layer neurons $\{X_1, X_2, \ldots, X_k\}$
2:  **Output:** Trained reduced model with aggregated outputs $[\chi_1, \chi_2, \ldots, \chi_m]$
3:  **Step 1: Representation of Neuron Outputs**
4:  Represent the outputs $\{X_1, X_2, \ldots, X_k\}$ as vectors in a high-dimensional space.
5:  **Step 2: Clustering Algorithm**
6:  Apply a clustering algorithm (e.g., $k$-means, hierarchical clustering) to group the outputs into $m$ clusters, where $m < k$.
7:  **for** each cluster $C_j$ where $j = 1$ to $m$ **do**
8:      Compute the aggregated output $\chi_j$ for cluster $C_j$.
9:      The aggregation could be the centroid, mean, sum, or maximum of the outputs in $C_j$.
10: **end for**
11: **Step 3: Model Adjustment**
12: Replace the original neuron outputs X with the new aggregated outputs $\chi$ in the network architecture.
13: The new model now uses $m$ neurons instead of $k$.
14: **Step 4: Training and Evaluation**
15: Train the reduced model using the training dataset.
16: Evaluate the models performance on validation and test datasets.
17: Compare the performance of the reduced model with the original model.
18: **Return** Trained reduced model

---

---

**Algorithm 4** Testing Neuron Output Efficiency

---
1:  **Input:** Outputs of Hidden neuron $i$ for all output nodes $\{Y_1, Y_2, \ldots, Y_q\}$
2:  **Output:** Test result for neuron $i$
3:  **Step 1: Data Collection**
4:  Gather the outputs $\{X_i^{(1)}, X_i^{(2)}, \ldots, X_i^{(q)}\}$ of neuron $i$ across all output nodes.
5:  **Step 2: Perform Statistical Test**
6:  Choose an appropriate statistical test (e.g., ANOVA, t-test) to compare the outputs $\{X_i^{(1)}, X_i^{(2)}, \ldots, X_i^{(q)}\}$.
7:  Compute the test statistic and $p$-value.
8:  **Step 3: Evaluate Test Statistic**
9:  Compare the $p$-value with the significance level $\alpha$.
10: **Decision Making**
11: **if** $p$-value $< \alpha$ **then**
12:     **Reject** $H_0$. **Conclusion:** There is significant variation in $X_i$s outputs across output nodes.
13: **else**
14:     **Fail to Reject** $H_0$. **Conclusion:** There is no significant variation in $X_i$s outputs.
15: **end if**
16: **Return** Test result for neuron $i$

---

---

**Algorithm 5** Neuron Reduction via PCA

---

1: **Input:** Outputs of hidden layer neurons $\{X_1, X_2, \ldots, X_k\}$
2: **Output:** Trained reduced model with principal components $\{Y_1, Y_2, \ldots, Y_m\}$
3: **Step 1: Representation of Neuron Outputs**
4: Represent the outputs $\{X_1, X_2, \ldots, X_k\}$ as vectors in a high-dimensional space.
5: **Step 2: Apply PCA**
6: Apply $PCA$ to the neuron outputs $\mathbf{X}$.
7: Retain the top $m$ principal components $\mathbf{Y} = [Y_1, Y_2, \ldots, Y_m]$, where $m < k$.
8: **Step 3: Model Adjustment**
9: Replace the original neuron outputs $\mathbf{X}$ with the new principal components $\mathbf{Y}$ in the network architecture.
10: The new model now uses $m$ neurons instead of $k$.
11: **Step 4: Training and Evaluation**
12: Train the reduced model using the training dataset.
13: Evaluate the models performance on validation and test datasets.
14: Compare the performance of the reduced model with the original model.
15: **Return** Trained reduced model with principal components $\{Y_1, Y_2, \ldots, Y_m\}$

---

---

**Algorithm 6** Model Accuracy CI Based on CLT

---

1: Input: Accuracy values $\{ACC_1, ACC_2, \ldots, ACC_n\}$, Confidence level
2: **Output:** CI [CLI, CUI]
3: Calculate the mean accuracy:

$$A\bar{C}C \leftarrow \frac{1}{n} \sum_{i=1}^{n} ACC_i$$

4: Calculate the standard deviation:

$$s \leftarrow \sqrt{\frac{1}{n-1} \sum_{i=1}^{n} (ACC_i - A\bar{C}C)^2}$$

5: Determine the significance level:

$$\alpha \leftarrow 1 - Confidence Level$$

6: Calculate the critical Z-value from the Z-table:

$$Z_{critical} \leftarrow Z-value at \left(1 - \frac{\alpha}{2}\right)$$

7: Compute the margin of error:

$$Margin of Error \leftarrow Z_{critical} \times \frac{s}{\sqrt{n}}$$

8: Compute the CLI:

$$CLI \leftarrow A\bar{C}C - Margin of Error$$

9: Compute the CUI:

$$CUI \leftarrow A\bar{C}C + Margin of Error$$

10: **Return:** [CLI, CUI]

---

---

**Algorithm 7** Model Accuracy CI Based on Bootstrap

---

1: **Input:** Training data $(X, Y)$, Number of bootstrap samples $B$, Initial model weights $\hat{\Theta}$, Confidence level $\alpha$
2: Output: CI $[CLI, CUI]$
3: **Step 1: Train Base Model**
4: Train a base model on $(X, Y)$ to obtain estimated parameters $\hat{\Theta}$
5: **Model:**

$$\hat{\Theta} = \arg\min_{\Theta} \frac{1}{n} \sum_{i=1}^{n} \mathcal{L}(f(x_i; \Theta), y_i)$$

6: **Step 2: Generate Bootstrap Samples**
7: **for** $i = 1\,to\,B$ **do**
8:     Generate bootstrap sample $(X_i^*, Y_i^*)$ by sampling $n$ data points with replacement from $(X, Y)$
9: **end for**
10: **Step 3: Train and Evaluate on Bootstrap Samples**
11: **for** $i = 1\,to\,B$ **do**
12:     Train a new model on bootstrap sample $(X_i^*, Y_i^*)$ using transfer learning with initial weights $\hat{\Theta}$
13:     Optimize model parameters $\Theta_i^*$ on $(X_i^*, Y_i^*)$
14:     Compute accuracy for $i$-th bootstrap sample:

$$ACC^{(i)} = \frac{1}{n} \sum_{j=1}^{n} \mathbb{I}\left(f(x_{ij}^*; \Theta_i^*) = y_{ij}^*\right)$$

15: **end for**
16: **Step 4: Compute Bootstrap Estimator**
17: Compute bootstrap mean accuracy:

$$A\bar{C}C = \frac{1}{B} \sum_{i=1}^{B} ACC^{(i)}$$

18: **Step 5: Compute Differences and Sort**
19: **for** $i = 1\,to\,B$ **do**
20:     Compute difference:

$$d_i = ACC^{(i)} - A\bar{C}C$$

21: **end for**
22: Sort differences $\{d_i\}$ in increasing order to get $d_{(1)}, d_{(2)}, \ldots, d_{(B)}$
23: **Step 6: Compute Percentiles**
24: Calculate lower percentile at $100\left(\frac{\alpha}{2}\right)\%$:

$$d_{(k)} = value\,at\,k = \left\lceil \frac{\alpha}{2} \cdot B \right\rceil$$

25: Calculate upper percentile at $100\left(1 - \frac{\alpha}{2}\right)\%$:

$$d_{(l)} = value\,at\,l = \left\lfloor \left(1 - \frac{\alpha}{2}\right) \cdot B \right\rfloor$$

26: **Step 7: Compute CI**
27: Compute lower limit of CI:

$$CLI = A\bar{C}C - d_{(l)}$$

28: Compute upper limit of CI:

$$CUI = A\bar{C}C - d_{(k)}$$

29: **Return:** CI $[CLI, CUI]$

---